\documentclass[conference]{IEEEtran}
\usepackage[utf8]{inputenc}
\usepackage{cite}\usepackage[colorlinks=true,pagebackref=false,urlcolor=black,linkcolor=black,citecolor=black]{hyperref}
\usepackage{balance}
\usepackage{graphicx}
\usepackage{booktabs}
\usepackage{url}
\usepackage[acronym]{glossaries}
\usepackage{multirow}
\usepackage{xcolor}
\usepackage[caption=false]{subfig}

\newacronym{alpr}{ALPR}{Automatic License Plate Recognition}
\newacronym{lp}{LP}{License Plate}
\newacronym{lpd}{LPD}{License Plate Detection}
\newacronym{vd}{VD}{Vehicle Detection}
\newacronym{cs}{CS}{Character Segmentation}
\newacronym{cr}{CR}{Character Recognition}
\newacronym{dataset}{UFPR-ALPR}{UFPR-ALPR}
\newacronym{ssig}{SSIG}{SSIG SegPlate Database}
\newacronym{png}{PNG}{Portable Network Graphics}
\newacronym{fps}{FPS}{Frames Per Second}
\newacronym{cnn}{CNN}{Convolutional Neural Network}
\newacronym{dl}{DL}{Deep Learning}
\newacronym{iou}{IoU}{Intersection over Union}
\newacronym{svm}{SVM}{Support Vector Machines}
\newacronym{hog}{HOG}{Histogram of Oriented Gradients}
\newacronym{rcnn}{RCNN}{Region-based CNN}
\newacronym{hmm}{HMM}{Hidden Markov Model}
\newacronym{snow}{SNoW}{Sparse Network of Winnows}
\newacronym{rnn}{RNN}{Recurrent Neural Network}
\newacronym{ctc}{CTC}{Connectionist Temporal Classification}

\begin{document}
\title{A Robust Real-Time Automatic License Plate Recognition Based on the YOLO Detector}

\author{
\IEEEauthorblockN{Rayson Laroca\IEEEauthorrefmark{1}, Evair Severo\IEEEauthorrefmark{1}, Luiz A. Zanlorensi\IEEEauthorrefmark{1}, Luiz S. Oliveira\IEEEauthorrefmark{1},\\Gabriel Resende Gon\c{c}alves\IEEEauthorrefmark{2}, William Robson Schwartz\IEEEauthorrefmark{2} and David Menotti\IEEEauthorrefmark{1}}
\IEEEauthorblockA{
\IEEEauthorrefmark{1}Department of Informatics, Federal University of Paran\'a (UFPR), Curitiba, PR, Brazil\, \\
\IEEEauthorrefmark{2}Department of Computer Science, Federal University of Minas Gerais (UFMG), Belo Horizonte, MG, Brazil\\
\\
\, Email: {\{rblsantos, ebsevero, lazjunior, lesoliveira, menotti\}}@inf.ufpr.br \quad \{gabrielrg, william\}@dcc.ufmg.br
}
}

\maketitle

\begin{abstract}
\gls*{alpr} has been a frequent topic of research due to many practical applications. 
However, many of the current solutions are still not robust in real-world situations, commonly depending on many constraints. 
This paper presents a robust and efficient \gls*{alpr} system based on the state-of-the-art YOLO object detector. 
The \glspl*{cnn} are trained and fine-tuned for each \gls*{alpr} stage so that they are robust under different conditions (e.g., variations in camera, lighting, and background). 
Specially for character segmentation and recognition, we design a two-stage approach employing simple data augmentation tricks such as inverted \glspl*{lp} and flipped characters. The resulting \gls{alpr} approach achieved impressive results in two datasets.
First, in the \acrshort*{ssig} dataset, composed of 2,000 frames from 101 vehicle videos, our system achieved a recognition rate of 93.53\% and 47 \gls*{fps}, performing better than both Sighthound and OpenALPR commercial systems (89.80\% and 93.03\%, respectively) and considerably outperforming previous results (81.80\%). Second, targeting a more realistic scenario,  we  introduce a larger public dataset\footnote{The \acrshort*{dataset} dataset is publicly available to the research community at \url{https://web.inf.ufpr.br/vri/databases/ufpr-alpr/} subject to privacy restrictions.}, called \acrshort*{dataset} dataset, designed to \gls*{alpr}. This dataset contains 150 videos and 4,500 frames captured when both camera and vehicles are moving and also contains different types of vehicles (cars, motorcycles, buses and trucks). 
In our proposed dataset, the trial versions of commercial systems achieved recognition rates below 70\%. On the other hand, our system performed better, with recognition rate of 78.33\% and 35 \gls*{fps}.
\end{abstract}

\IEEEpeerreviewmaketitle

\section{Introduction}
\label{sec:intro}

\glsresetall

\gls*{alpr} has been a frequent topic of research~\cite{review2013,gou2016,bulan2017} due to many practical applications, such as automatic toll collection, traffic law enforcement, private spaces access control and road traffic monitoring. 

\gls*{alpr} systems typically have three stages: \gls*{lp} detection, character segmentation and character recognition. The earlier stages require higher accuracy or almost perfection, since failing to detect the \gls*{lp} would probably lead to a failure in the next stages either. Many approaches search first for the vehicle and then its \gls*{lp} in order to reduce processing time and eliminate false positives.

Although \gls*{alpr} has been frequently addressed in the literature, many studies and solutions are still not robust enough on real-world scenarios. These solutions commonly depend on certain constraints, such as specific cameras or viewing angles, simple backgrounds, good lighting conditions, search in a fixed region, and certain types of vehicles (they would not detect \glspl*{lp} from vehicles such as motorcycles, trucks or buses).

Many computer vision tasks have recently achieved a great increase in performance mainly due to the availability of large-scale annotated datasets (i.e., ImageNet~\cite{imagenet2009}) and the hardware (GPUs) capable of handling a large amount of data. In this scenario, \gls*{dl} techniques arise. However, despite the remarkable progress of \gls*{dl} approaches in \gls*{alpr}~\cite{li2016reading,li2017towards, masood2017sighthound}, there is still a great demand for \gls*{alpr} datasets with vehicles and \glspl*{lp} annotations. The amount of training data is determinant for the performance of \gls*{dl} techniques. 
Higher amounts of data allow the use of more robust network architectures with more parameters and layers.
Hence, we propose a larger benchmark dataset, called \acrshort*{dataset}, focused on different real-world scenarios.

To the best of our knowledge, the \gls*{ssig}~\cite{goncalves2016benchmark} is the largest public dataset of Brazilian \glspl*{lp}. This dataset contains less than $800$ training examples and has several constraints such as: it uses a static camera mounted always in the same position, all images have very similar and relatively simple backgrounds, there are no motorcycles and only a few cases where the \glspl*{lp} are not well aligned.
	
When recording the \acrshort*{dataset} dataset, we sought to eliminate many of the constraints found in \gls*{alpr} applications by using three different non-static cameras to capture $4$,$500$ images from different types of vehicles (cars, motorcycles, buses, trucks, among others) with complex backgrounds and under different lighting conditions. 
The vehicles are in different positions and distances to the camera. 
Furthermore, in some cases, the vehicle is not fully visible on the image. 
To the best of our knowledge, there are no public datasets for \gls*{alpr} with annotations of cars, motorcycles, \glspl*{lp} and characters.
Therefore, we can point out two main challenges in our dataset.
First, usually, car and motorcycle \glspl*{lp} have different aspect ratios, not allowing \gls*{alpr} approaches to use this constraint to filter false positives.
Also car and motorcycle \glspl*{lp} have different layouts and positions.

As great advances in object detection were achieved through YOLO-inspired models~\cite{ning2017spatially,wu2016squeeze}, we decided to fine-tune it for \gls*{alpr}.
YOLOv2~\cite{redmon2016yolo9000} is a state-of-the-art real-time object detection that uses a model with $19$ convolutional layers and $5$ maxpooling layers. On the other hand, Fast-YOLO~\cite{redmon2016yolo} is a model focused on a speed/accuracy trade-off that uses fewer convolutional layers ($9$ instead of $19$) and fewer filters in those layers. Therefore, Fast-YOLO is much faster but less accurate than YOLOv2.

In this work, we propose a new robust real-time \gls*{alpr} system based on the YOLO object detection \glspl*{cnn}.
Since we are processing video frames, we also employ temporal redundancy such that we process each frame independently and then combine the results to create a more robust prediction for each vehicle. 

The proposed system outperforms previous results and two commercial systems in the \gls*{ssig} dataset and also in our proposed \acrshort*{dataset}.
The main contributions of this paper can be summarized as follows: 
\begin{itemize}
	\item  A new real-time end-to-end \gls*{alpr} system using the state-of-the-art YOLO object detection \glspl*{cnn}\footnote{The entire \gls*{alpr} system, i.e., the architectures and weights, is publicly available for academic purposes.};
	\item A robust two-stage approach for character segmentation and recognition mainly due to simple data augmentation tricks for training data such as inverted \glspl*{lp} and flipped characters.
	\item A  public dataset for \gls*{alpr} with $4$,$500$ fully annotated images (over $30$,$000$ \gls*{lp} characters) focused on \textbf{usual and different} real-world scenarios, showing that our proposed \gls*{alpr} system yields outstanding results in both scenarios. 
	\item A comparative evaluation among the proposed approach, previous works in the literature and two commercial systems in the \acrshort*{dataset} dataset.
\end{itemize}

This paper is organized as follows. 
We briefly review related work in Section~\ref{sec:related_work}. The \acrshort*{dataset} dataset is introduced in Section~\ref{sec:dataset}. Section~\ref{sec:proposed_system} presents the proposed \gls*{alpr} system using object detection \glspl*{cnn}. We report and discuss the results of our experiments in Section~\ref{sec:results}. Conclusions and future work are given in Section~\ref{sec:conclusion}.

\section{Related Work}
\label{sec:related_work}

In this section, we briefly review several recent works that use \gls*{dl} approaches in the context of \gls*{alpr}. 
For relevant studies using conventional image processing techniques, please refer to~\cite{anagnostopoulos2008,hsu2013application,ashtari2014iranian,sarfraz2013real,review2013,panahi2017accurate,gou2016,yuan2017,azam2016automatic}. 
More specifically, we discuss works related to each \gls*{alpr} stage, and specially studies works that not fit into the other subsections.
This section concludes with final remarks.

\noindent \textbf{\textit{\gls*{lp} Detection}}:
Many authors have addressed the \gls*{lp} detection stage with object detection \glspl*{cnn}. Montazzolli and Jung~\cite{montazzolli2017} used a single \gls*{cnn} arranged in a cascaded manner to detect both car frontal-views and its \glspl*{lp}, achieving high
recall and precision rates. Hsu et al.~\cite{hsu2017robust} customized \glspl*{cnn} exclusively for \gls*{lp} detection and demonstrated that the modified versions perform better. Rafique et al.~\cite{rafique2017vehicle} applied \gls*{svm} and \gls*{rcnn} for \gls*{lp} detection, noting that \glspl*{rcnn} are best suited for real-time systems.

Li and Chen~\cite{li2016reading} trained a \gls*{cnn} based on characters cropped from general text to perform a character-based \gls*{lp} detection, achieving higher recall and precision rates than previous approaches. Bulan et al.~\cite{bulan2017} first extracts a set of candidate \gls*{lp} regions using a weak \gls*{snow} classifier and then filters them using a strong \gls*{cnn}, significantly improving the baseline method.

\noindent \textbf{\textit{Character Segmentation}}:
\gls*{alpr} systems based on \gls*{dl} techniques usually address the character segmentation and recognition together. 
Montazzolli and Jung~\cite{montazzolli2017} propose a \gls*{cnn} to segment and recognize the characters within a cropped \gls*{lp}.
They have segmented more than $99\%$ of the characters correctly, outperforming the baseline by a large margin.

Bulan et~al.~\cite{bulan2017} achieved very high accuracy in \gls*{lp} recognition jointly performing the character segmentation and recognition using \glspl*{hmm} where the most likely \gls*{lp} was determined by applying the Viterbi algorithm.

\noindent \textbf{\textit{Character Recognition}}:
Menotti et al.~\cite{menotti2014vehicle} proposed the use of random \glspl*{cnn} to extract features for character recognition, achieving a significantly better performance than using image pixels or learning the filters weights with back-propagation. Li and Chen~\cite{li2016reading} proposed to perform the character recognition as a sequence labelling problem. A \gls*{rnn} with \gls*{ctc} is employed to label the sequential data, recognizing the whole \gls*{lp} without the character-level segmentation.

Although Svoboda et al.~\cite{svoboda2016cnn} have not perform the character recognition itself, they achieved high quality \gls*{lp} deblurring reconstructions using a text deblurring \gls*{cnn}, which can be very useful in character recognition. 

\noindent \textbf{\textit{Miscellaneous}}:
Masood et al.~\cite{masood2017sighthound} presented an end-to-end \gls*{alpr} system using a sequence of deep \glspl*{cnn}. As this is a commercial system, little information is given about the used \glspl*{cnn}. 
Li et al.~\cite{li2017towards} propose a unified \gls*{cnn} that can locate \glspl*{lp} and recognize them simultaneously in a single forward pass. In addition, the model size is highly decreased by sharing many of its convolutional features. 

\noindent \textbf{\textit{Final Remarks}}:
Many papers only address part of the \gls*{alpr} pipeline (e.g., \gls*{lp} detection) or perform their experiments on datasets that do not represent real-world scenarios, making it difficult to accurately evaluate the presented methods. In addition, most of the approaches are not capable of recognizing \glspl*{lp} in real-time, making it impossible for them to be applied in some applications. In this sense, we employ the YOLO object detection \glspl*{cnn} in each stage to create a robust and efficient end-to-end \gls*{alpr} system. In addition,  we perform data augmentation for character recognition, since this stage is the bottleneck in some \gls*{alpr} systems.
\section{The UFPR-ALPR Dataset}
\label{sec:dataset}

\begin{figure*}[!htb]
	\centering
	\includegraphics[width=0.24\textwidth]{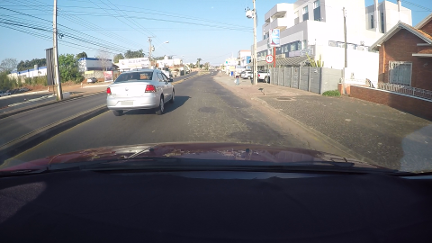}
	\includegraphics[width=0.24\textwidth]{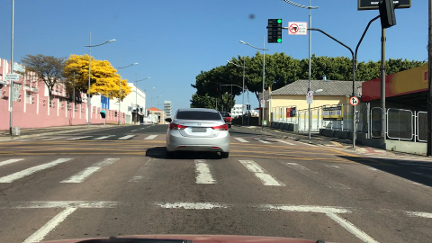} \vspace{0.1cm}
	\includegraphics[width=0.24\textwidth]{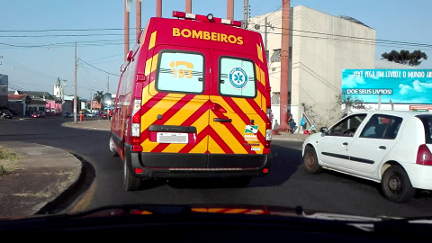}
	\includegraphics[width=0.24\textwidth]{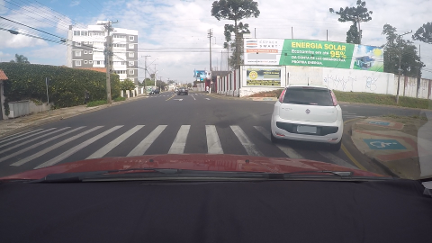} \vspace{0.1cm}
	\includegraphics[width=0.24\textwidth]{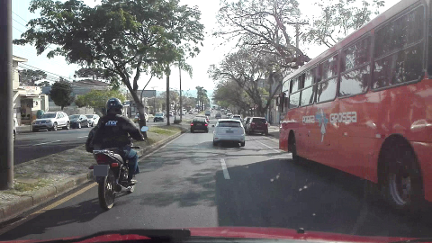}
	\includegraphics[width=0.24\textwidth]{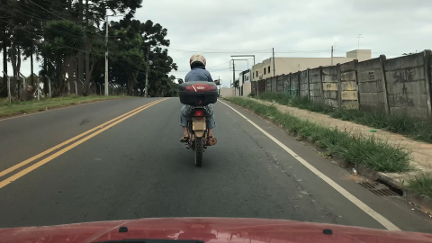}
	\includegraphics[width=0.24\textwidth]{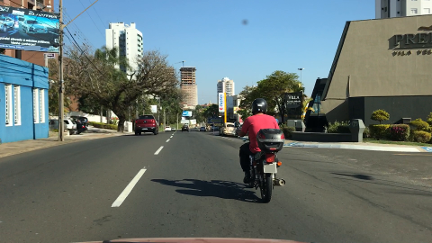} 
	\includegraphics[width=0.24\textwidth]{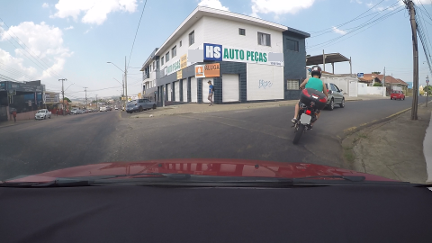} \vspace{0.1cm}
	\includegraphics[width=0.24\textwidth]{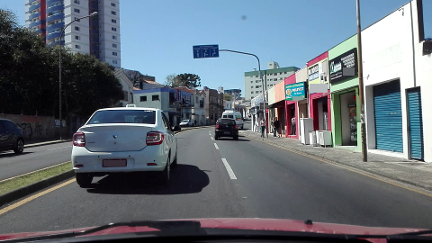}
	\includegraphics[width=0.24\textwidth]{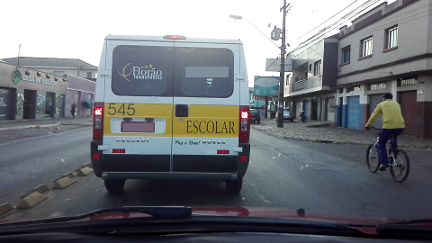}
	\includegraphics[width=0.24\textwidth]{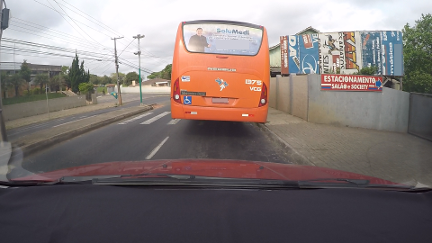}
	\includegraphics[width=0.24\textwidth]{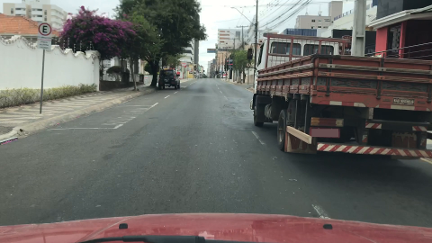}
	\includegraphics[width=0.24\textwidth]{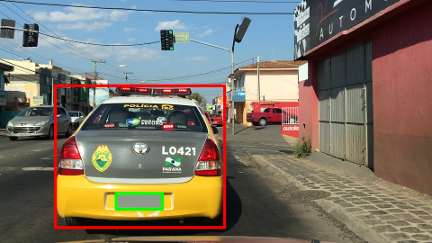}
	\includegraphics[width=0.24\textwidth]{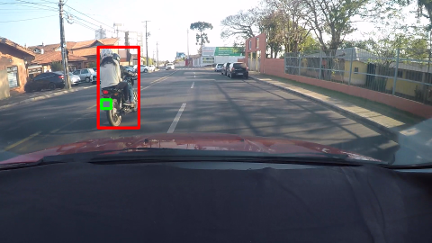}
	\includegraphics[width=0.24\textwidth]{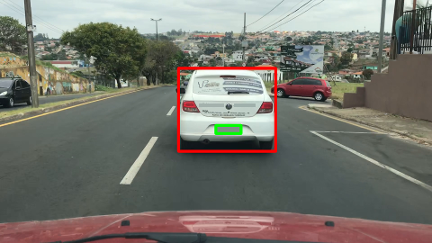}
	\includegraphics[width=0.24\textwidth]{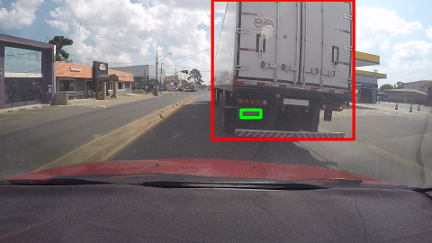}
	\caption{Sample images of the \acrshort*{dataset} dataset. First three rows show the variety in backgrounds, lighting conditions, as well as vehicle/\gls*{lp} positions and types. Fourth row shows examples of vehicle and \gls*{lp} annotations. The \glspl*{lp} were blurred due to privacy constraints.}
	\label{fig:dataset}    
\end{figure*}

The dataset contains $4$,$500$ images taken from inside a vehicle driving through regular traffic in an urban environment. These images were obtained from $150$ videos with duration of $1$ second and frame rate of $30$ \gls*{fps}.
Thus, the dataset is divided into $150$ vehicles, each with $30$ images with only one visible \gls*{lp} in the foreground. It is noteworthy that no stabilization method was used. Fig.~\ref{fig:dataset} shows the diversity of the dataset.

The images were acquired with three different cameras and are available in the \gls*{png} format with size of $1$,$920$ $\times$ $1$,$080$ pixels. The cameras used were:~\emph{GoPro Hero4 Silver}, \emph{Huawei P9 Lite} and \emph{iPhone 7 Plus}. Images obtained with different cameras do not necessarily have the same quality, although they have the same resolution and frame rate. This is due to different camera specifications, such as autofocus, bit rate, focal length and optical image stabilization.

There are minor variations in the camera position due to repeated mountings of the camera and also to simulate a real condition, where the camera is not always placed in exactly the same position.

\begin{figure}[!htb]
	\centering
	\subfloat[][Car \glspl*{lp}\label{fig:car_plates}]{%
		\includegraphics[]{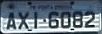}
		\includegraphics[]{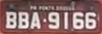}}
		
    \subfloat[][Motorcycle \glspl*{lp}\label{fig:motorcycle_plates}]{%
		\includegraphics[]{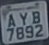}
		\includegraphics[]{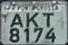}
		\includegraphics[]{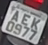}}
	\caption{Examples of the different \gls*{lp} types found in the \acrshort*{dataset} dataset. In Brazil, cars'~\glspl*{lp} have $3$~letters and $4$~digits in the same row and motorcycles'~\glspl*{lp} have $3$~letters in one row and $4$~digits in another.} 
	\label{fig:plates}    
\end{figure}

We collected $1$,$500$ images with each camera, divided as follows: $900$ of cars with gray \gls*{lp}, $300$ of cars with red \gls*{lp} and $300$ of motorcycles with gray \gls*{lp}. In Brazil, the \glspl*{lp} have size and color variations depending on the type of the vehicle and its category. Cars'~\glspl*{lp} have a size of $40$cm~$\times$~$13$cm, while motorcycles~\glspl*{lp} have $20$cm~$\times$~$17$cm. Private vehicles have gray \glspl*{lp}, while buses, taxis and other transportation vehicles have red \glspl*{lp}. There are other color variations for specific categories such as official or older cars. Fig.~\ref{fig:plates} shows some of the different types of \glspl*{lp} found in the dataset. 

The dataset is split as follows: $40\%$ for training, $40\%$~for testing and $20\%$ for validation, using the same protocol division proposed by Gon\c{c}alves et al.~\cite{goncalves2016benchmark} in the \gls*{ssig} dataset. 
The dataset distribution was made so that each split has the same number of images obtained with each camera, taking into account the type and position of the vehicle, the color and the characters of the vehicle's \gls*{lp}, the distance of the vehicle from the camera (based on the height of the \gls*{lp} in pixels) such that each split is as representative as possible.

\begin{figure}[!htb]
	\centering
	\vspace{-12pt}
	\subfloat[][Vehicles (\gls*{ssig})\label{fig:heat_v_ssig}]{
		\includegraphics[width=0.48\columnwidth]{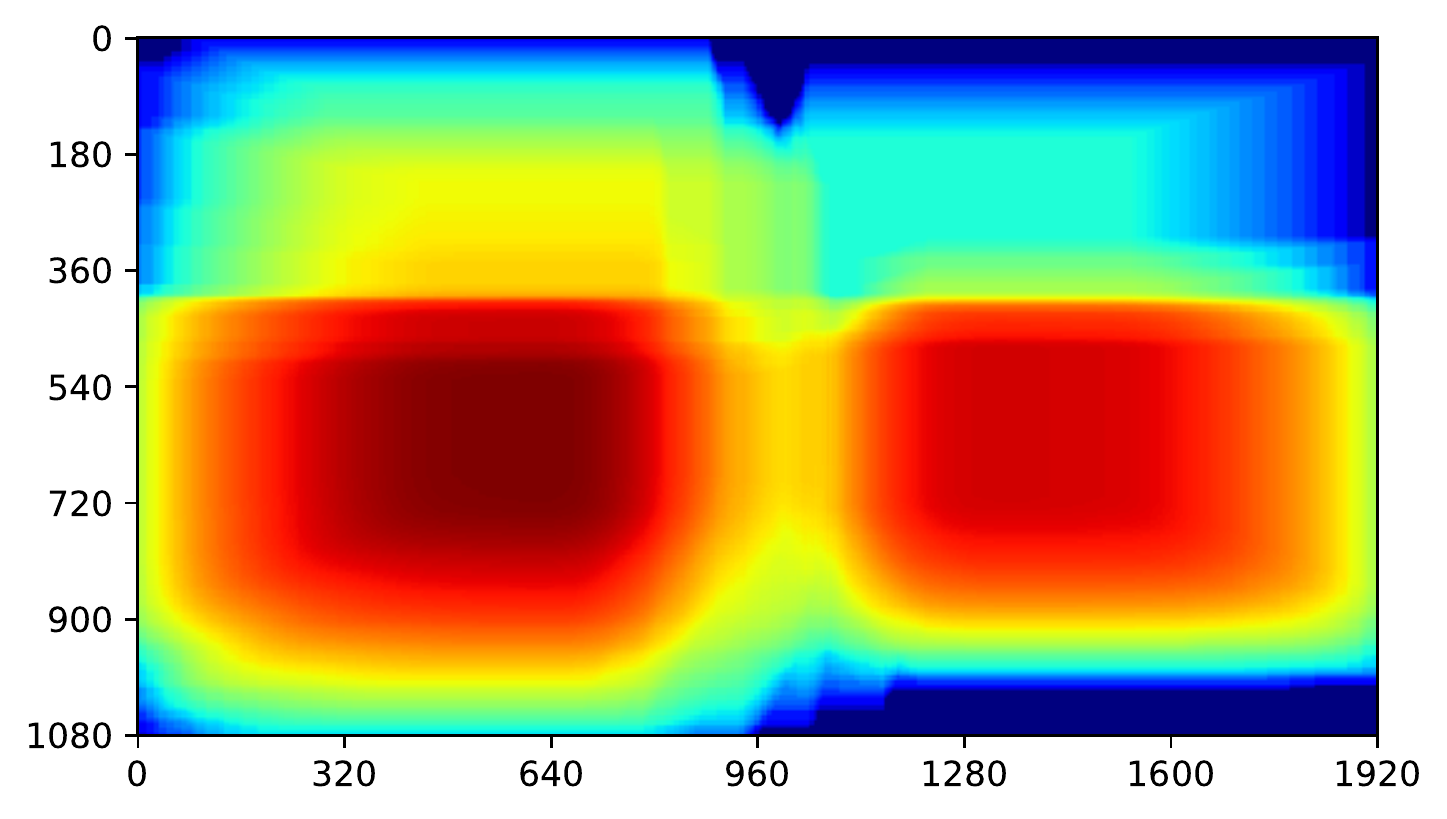}}%
    \subfloat[][\glspl*{lp} (\gls*{ssig})\label{fig:heat_lp_ssig}	]{
		\includegraphics[width=0.48\columnwidth]{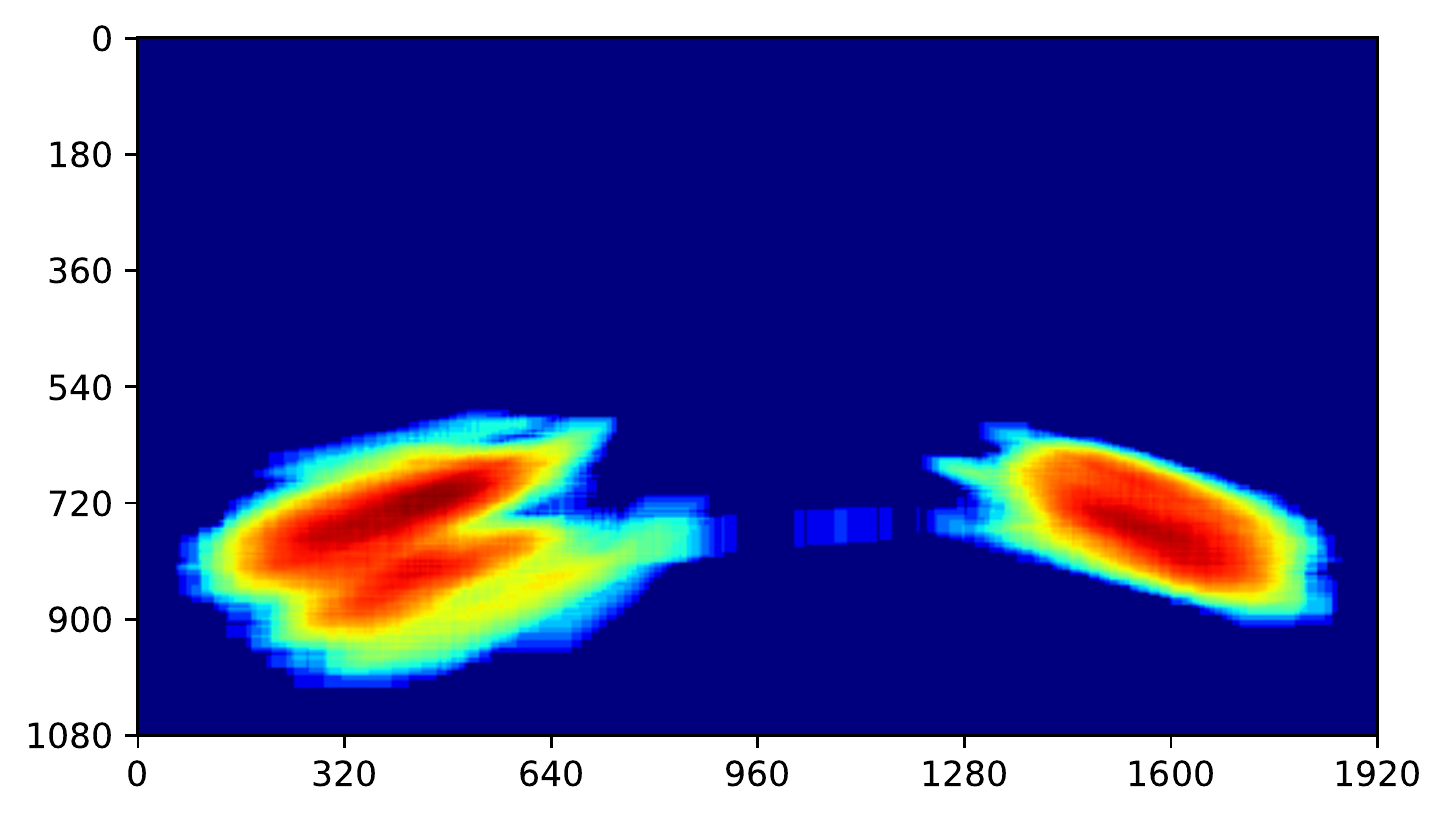}} \\[-1ex]
		
	\subfloat[][Vehicles (\acrshort*{dataset})\label{fig:heat_v}]{
		\includegraphics[width=0.48\columnwidth]{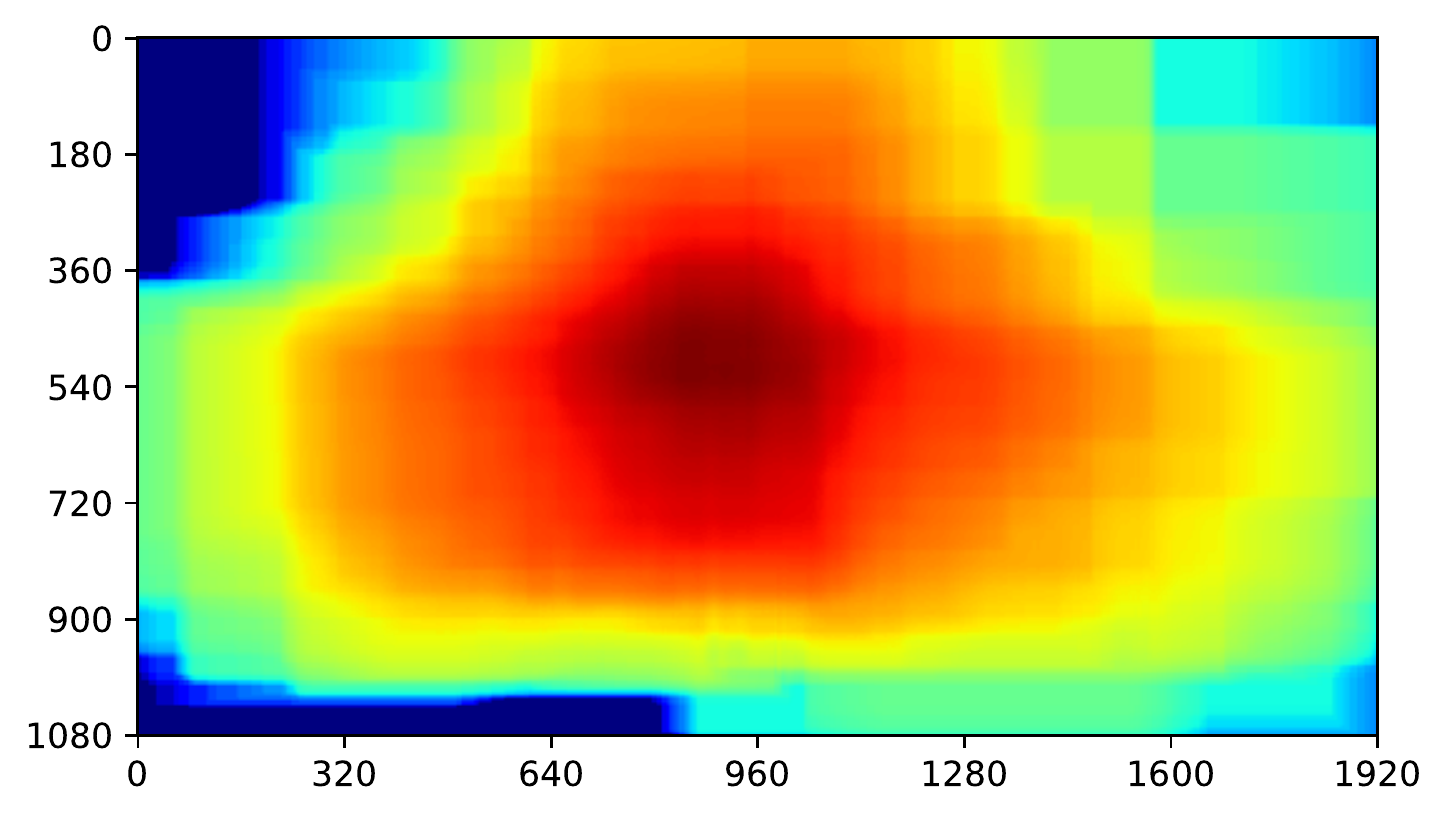}}%
    \subfloat[][\glspl*{lp} (\acrshort*{dataset})\label{fig:heat_lp}]{
		\includegraphics[width=0.48\columnwidth]{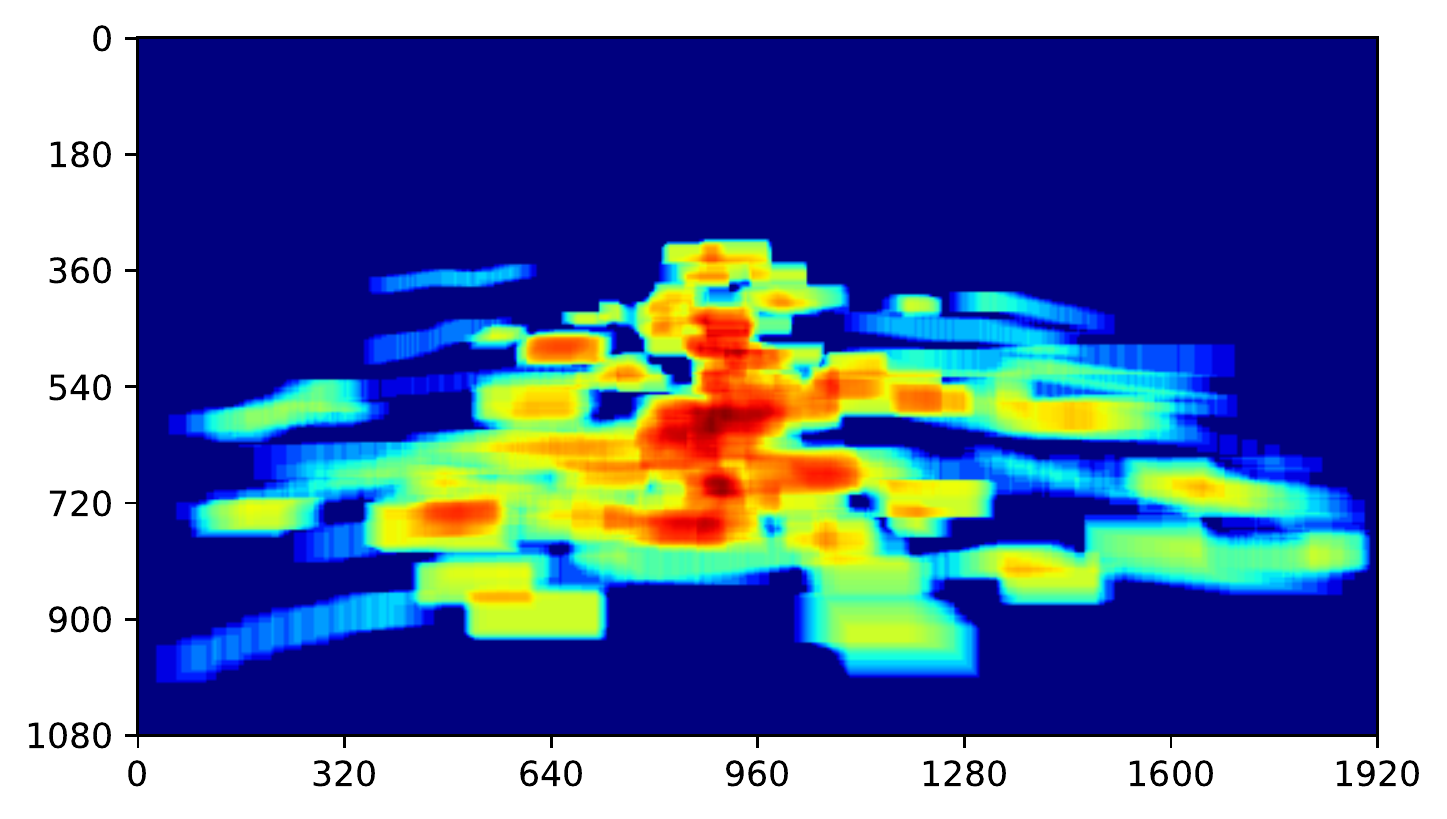}}
	\caption{Heat maps illustrating the distribution of vehicles and \glspl*{lp} in the \gls*{ssig} and \gls*{dataset} datasets. The heat maps are log-normalized, meaning the distribution is even more concentrated than it appears.}
	\label{fig:heat_maps}
	\vspace{-2mm} 
\end{figure}

The heat maps of the distribution of the vehicles and \glspl*{lp} for the image frame in both \gls*{ssig} and \acrshort*{dataset} datasets are shown in Fig.~\ref{fig:heat_maps}. 
As can be seen, the vehicles and \glspl*{lp} are much better distributed in our dataset.

\begin{figure}[!htb]
	\centering
 	\includegraphics[width=\columnwidth]{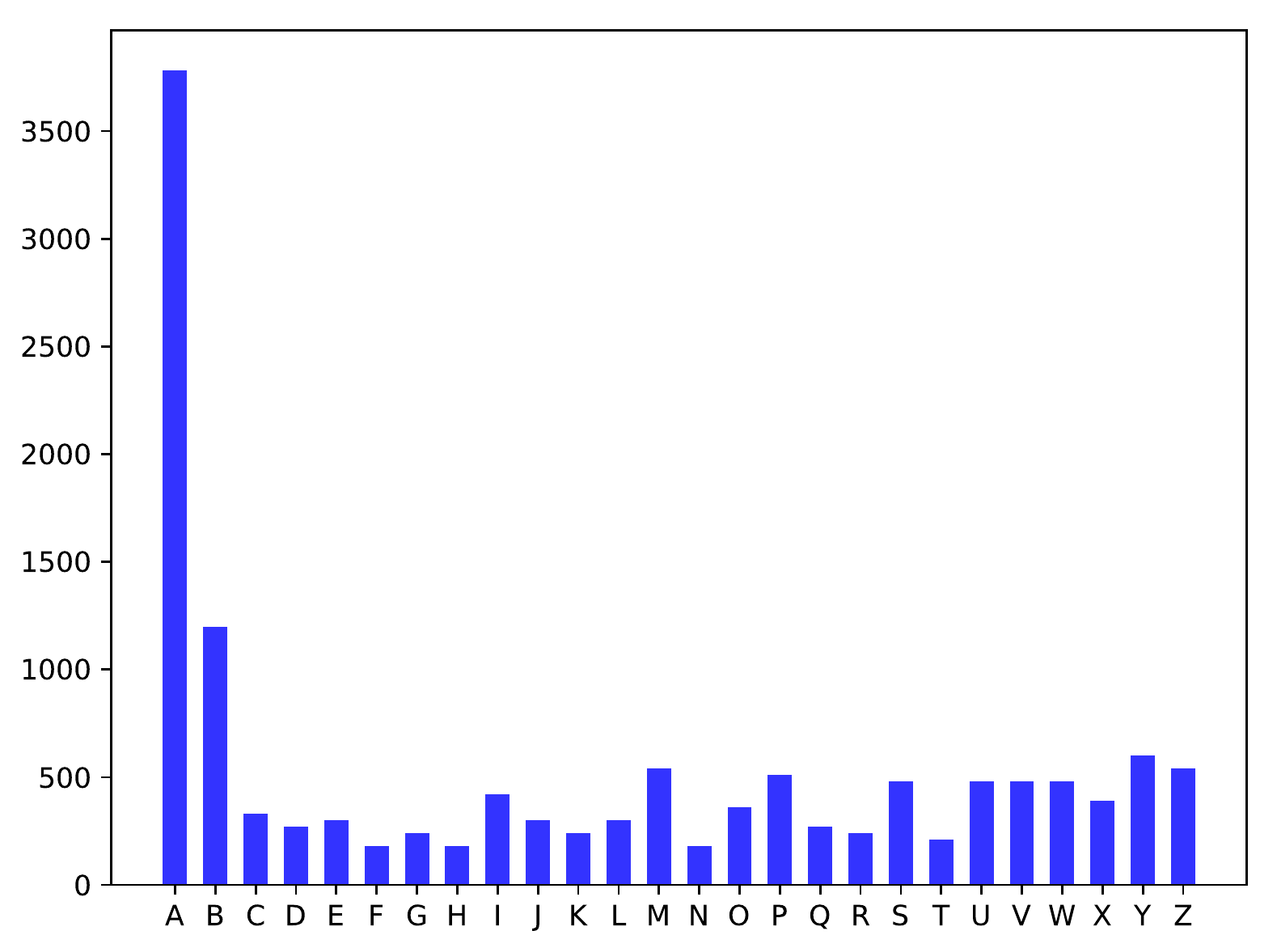}
 	\vspace*{-7mm}
	\caption{Letters distribution in the \gls*{dataset} dataset.}
	\label{fig:letters_frequency}    
\end{figure}

In Brazil, each state uses particular starting letters for its \glspl*{lp} which results in a specific range. In Paran\'a (where the dataset was collected), \glspl*{lp} range from AAA-0001 to BEZ-9999. Therefore, the letters A and B have many more examples than the others, as shown in Fig.~\ref{fig:letters_frequency}.

Every image has the following annotations available in a text file: the camera in which the image was taken, the vehicle's position and information such as: type (car or motorcycle), manufacturer, model and year; the identification and position of the \gls*{lp}, as well as the position of its characters. Fig.~\ref{fig:dataset} shows the bounding boxes of different types of vehicle and \glspl*{lp}. 
\section{Proposed ALPR Approach}
\label{sec:proposed_system}

\begin{figure*}[!htb]
	\centering
	\includegraphics[width=\textwidth]{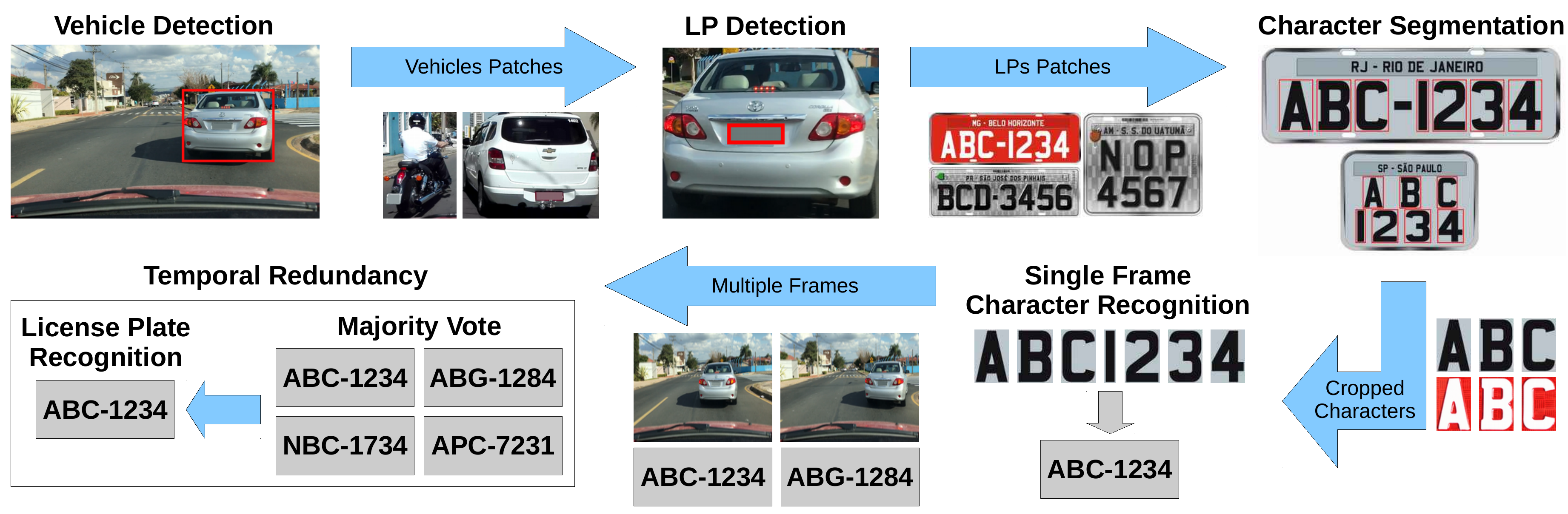}
	\caption{An usual \gls*{alpr} pipeline having temporal redundancy at the end.}
	\label{fig:alpr_pipeline}    
\end{figure*}

This section describes the proposed approach and it is divided into four subsections, one for each of the \gls*{alpr} stages (i.e., vehicle and \gls*{lp} detection, character segmentation and character recognition) and one for temporal redundancy. Fig.~\ref{fig:alpr_pipeline} illustrates the \gls*{alpr} pipeline, explained throughout this section.

We use specific \glspl*{cnn} for each \gls*{alpr} stage. Thus, we can tune the parameters separately in order to improve the performance for each task. The models used are: Fast-YOLO, YOLOv2 and CR-NET~\cite{montazzolli2017}, an architecture inspired by Fast-YOLO for character segmentation and recognition.

\subsection{Vehicle and LP Detection}
\label{system:vehicle_detection}

We train two \glspl*{cnn} in this stage: one for vehicle detection in the input image and other for \gls*{lp} detection in the detected vehicle. Recent works~\cite{montazzolli2017,goncalves2016} also performed the vehicle detection first.

We evaluated both Fast-YOLO and YOLOv2 models at this stage to be able to handle simpler (i.e.,~\gls*{ssig}) and more realistic (i.e.,~\acrshort*{dataset}) data. For simpler scenarios, the Fast-YOLO should be able to detect the vehicles and their \glspl*{lp} correctly in much less time. However, for more realistic scenarios it might not be deep enough to perform these tasks.

In order to use both YOLO models\footnote{For training YOLOv2 and Fast-YOLO we used convolutional weights pre-trained on ImageNet~\cite{imagenet2009}, available at \url{https://pjreddie.com/darknet/yolo/}.}, we need to change the number of filters in the last convolutional layer to match the number of classes. YOLO uses $A$ anchor boxes to predict bounding boxes (we use $A$ = $5$) each with four coordinates $(x, y, w, h)$, confidence and $C$ class probabilities~\cite{redmon2016yolo9000}, so the number of filters is given by 
\vspace{-1mm}
\begin{equation} 
\label{eq:filters}
filters = (C + 5) \times A.
\end{equation} 

In a dataset such as the \gls*{ssig} dataset, we intend to detect only one class in both vehicle and \gls*{lp} detection (first the car and then its \gls*{lp}), so the number of filters in each task has been reduced to $30$. On the other hand, the \acrshort*{dataset} dataset includes images from cars and motorcycles (two classes), so the number of filters in the vehicle detection task must be $35$. In our tests, the results were better when using two classes (instead of just one class called \lq{vehicle}\rq{}). The Fast-YOLO's architecture used in both tasks is shown in~Table~\ref{tab:fast_yolo}. The same changes were made in the YOLOv2 model architecture (not shown due to lack of space).

\begin{table}[!htb]
\vspace{-2.5mm}
\centering
\caption{Fast-YOLO network used in both vehicle and \gls*{lp} detection. There are either $30$ or $35$ filters in the last convolutional layer to detect one or two classes, respectively.}
\label{tab:fast_yolo}
\resizebox{\columnwidth}{!}{
\begin{tabular}{@{}cccccc@{}}
\toprule
\multicolumn{2}{c}{\textbf{Layer}} & \textbf{Filters} & \textbf{Size} & \textbf{Input} & \textbf{Output} \\ \midrule
$0$ & conv & $16$ & $3 \times 3 / 1$ & $416 \times 416 \times 3$ & $416 \times 416 \times 16$ \\
$1$ & max &  & $2 \times 2 / 2$ & $416 \times 416 \times 16$ & $208 \times 208 \times 16$ \\
$2$ & conv & $32$ & $3 \times 3 / 1$ & $208 \times 208 \times 16$ & $208 \times 208 \times 32$ \\
$3$ & max &  & $2 \times 2 / 2$ & $208 \times 208 \times 32$ & $104 \times 104 \times 32$ \\
$4$ & conv & $64$ & $3 \times 3 / 1$ & $104 \times 104 \times 32$ & $104 \times 104 \times 64$ \\
$5$ & max &  & $2 \times 2 / 2$ & $104 \times 104 \times 64$ & $52 \times 52 \times 64$ \\
$6$ & conv & $128$ & $3 \times 3 / 1$ & $52\times 52 \times 64$ & $52 \times 52 \times 128$ \\
$7$ & max &  & $2 \times 2 / 2$ & $52 \times 52 \times 128$ & $26 \times 26 \times 128$ \\
$8$ & conv & $256$ & $3 \times 3 / 1$ & $26 \times 26 \times 128$ & $26 \times 26 \times 256$ \\
$9$ & max &  & $2 \times 2 / 2$ & $26 \times 26 \times 256$ & $13 \times 13 \times 256$ \\
$10$ & conv & $512$ & $3 \times 3 / 1$ & $13 \times 13 \times 256$ & $13 \times 13 \times 512$ \\
$11$ & max &  & $2 \times 2 / 1$ & $13 \times 13 \times 512$ & $13 \times 13 \times 512$ \\
$12$ & conv & $1024$ & $3 \times 3 / 1$ & $13 \times 13 \times 512$ & $13 \times 13 \times 1024$ \\
$13$ & conv & $1024$ & $3 \times 3 / 1$ & $13 \times 13 \times 1024$ & $13 \times 13 \times 1024$ \\
$14$ & conv & $30/35$ & $1 \times 1 / 1$ & $13 \times 13 \times 1024$ & $13 \times 13 \times 30/35$ \\
$15$ & detection &  &  &  &  \\ \bottomrule
\end{tabular}}
\end{table}

While the entire frame and the vehicle coordinates are used as inputs to train the vehicle detection \gls*{cnn}, the vehicle patch (with a margin) and the coordinates of its \gls*{lp} are used to learn the \gls*{lp} detection network.
The size of the margin is defined as follows. We evaluated, in the validation set, the required margin so that all \glspl*{lp} would be completely within the bounding boxes of the vehicles found by the vehicle detection \gls*{cnn}. This is done to avoid losing \glspl*{lp} in cases where the vehicle is not very well detected/segmented.

By default, YOLO only returns objects detected with a confidence of $0.25$ or higher. 
In the validation set, we evaluated the best threshold in order to detect all vehicles having the lowest false positive rate.
A negative recognition result is given in cases where no vehicle is found. For \gls*{lp} detection we use threshold equal $0$, as there might be cases where the \gls*{lp} is detected with very low confidence (e.g., $0.1$). We keep only the detection with the largest confidence in cases where more than one \gls*{lp} is detected, since each vehicle has only one \gls*{lp}.

\subsection{Character Segmentation}
\label{system:cs}

Once the \gls*{lp} has been detected, we employ the \gls*{cnn} proposed by Montazzolli and Jung~\cite{montazzolli2017} (CR-NET) for character segmentation and recognition. However, instead of performing both stages at the same time through an architecture with $35$ classes (0-9, A-Z, where the letter~O is detected jointly with the digit~0), we chose to first use a network to segment the characters and then another two to recognize them. Knowing that all Brazilian \glspl*{lp} have the same format: three letters and four digits, we use $26$~classes for letters and $10$~classes for digits. As pointed out by Gonçalves et al.~\cite{goncalves2016}, this reduces the incorrect classification.

The character segmentation \gls*{cnn} (architecture described in Table~\ref{tab:yolo_montazzolli}) is trained using the \gls*{lp} patch (with a margin) and the characters coordinates as inputs. As in the previous stage, this margin is defined based on the validation set to ensure that all characters are completely within its predicted~\gls*{lp}.

\begin{table}[!htb]
\centering
\caption{Character segmentation \gls*{cnn}, proposed in~\cite{montazzolli2017}. We changed the number of filters in the last convolutional layer to $30$, as we want to first segment the character (one class).}
\label{tab:yolo_montazzolli}
\resizebox{\columnwidth}{!}{
\begin{tabular}{@{}cccccc@{}}
\toprule
\multicolumn{2}{c}{\textbf{Layer}} & \textbf{Filters} & \textbf{Size} & \textbf{Input} & \textbf{Output} \\ \midrule
$1$ & conv & $32$ & $3 \times 3 / 1$ & $240 \times 80 \times 3$ & $240 \times 80 \times 32$ \\
$2$ & max &  & $2 \times 2 / 2$ & $240 \times 80 \times 32$ & $120 \times 40 \times 32$ \\
$3$ & conv & $64$ & $3 \times 3 / 1$ & $120 \times 40 \times 32$ & $120 \times 40 \times 64$ \\
$4$ & max &  & $2 \times 2 / 2$ & $120 \times 40 \times 64$ & $60 \times 20 \times 64$ \\
$5$ & conv & $128$ & $3 \times 3 / 1$ & $60 \times 20 \times 64$ & $60 \times 20 \times 128$ \\
$6$ & conv & $64$ & $1 \times 1 / 1$ & $60 \times 20 \times 128$ & $60 \times 20 \times 64$ \\
$7$ & conv & $128$ & $3 \times 3 / 1$ & $60\times 20 \times 64$ & $60 \times 20 \times 128$ \\
$8$ & max & & $2 \times 2 / 2$ & $60 \times 20 \times 128$ & $30 \times 10 \times 128$ \\
$9$ & conv & $256$ & $3 \times 3 / 1$ & $30 \times 10 \times 128$ & $30 \times 10 \times 256$ \\
$10$ & conv & $128$ & $1 \times 1 / 1$ & $30 \times 10 \times 256$ & $30 \times 10 \times 128$ \\
$11$ & conv & $256$ & $3 \times 3 / 1$ & $30 \times 10 \times 128$ & $30 \times 10 \times 256$ \\
$12$ & conv & $512$ & $3 \times 3 / 1$ & $30 \times 10 \times 256$ & $30 \times 10 \times 512$ \\
$13$ & conv & $256$ & $1 \times 1 / 1$ & $30 \times 10 \times 512$ & $30 \times 10 \times 256$ \\
$14$ & conv & $512$ & $3 \times 3 / 1$ & $30 \times 10 \times 256$ & $30 \times 10 \times 512$ \\
$15$ & conv & $30$ & $1 \times 1 / 1$ & $30 \times 10 \times 512$ & $30 \times 10 \times 30$ \\
$16$ & detection &  &  &  &  \\ \bottomrule
\end{tabular}}
\end{table}

The \gls*{cnn} input size ($240$ $\times$ $80$) was chosen based on the \gls*{lp}'s ratio of Brazilian cars ($3$ $\times$ $1$), however the motorcycles \glspl*{lp} are nearly square ($1.17$ $\times$ $1$). That way, we enlarged horizontally all detected \glspl*{lp} (to $2.75$ $\times$ $1$) before performing the character segmentation. 

We also create a negative image of each \gls*{lp}, thereby doubling the number of training samples. Since the color of the characters in the Brazilian \glspl*{lp} depends on the category of the vehicle (e.g., private or commercial), the negative images simulate characters from other categories.

In some cases, more than $7$ characters might be detected. If there are no overlaps (\gls*{iou}~$\geq$~$0.25$), we discard the ones with the lowest confidence levels. Otherwise, we perform the union between the overlapping characters, turning them into a single character. As motorcycle \glspl*{lp} can be very tilted, we use a higher threshold (\gls*{iou}~$\geq$~$0.75$) to consider the overlap between its characters.

\subsection{Character Recognition}
\label{system:cr}

\begin{figure}[!htb]
	\vspace{-5mm} 
	\centering
	\captionsetup[subfigure]{labelformat=empty} 
	\subfloat[][$0$p]{%
		\includegraphics[width=5mm,height=8mm]{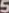}} \quad
	\subfloat[][$1$p]{%
		\includegraphics[width=5mm,height=8mm]{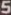}} \quad
	\subfloat[][$2$p]{%
		\includegraphics[width=5mm,height=8mm]{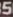}} \quad
	\subfloat[][$3$p]{%
		\includegraphics[width=5mm,height=8mm]{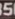}} \quad
	\subfloat[][$4$p]{%
		\includegraphics[width=5mm,height=8mm]{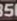}} \quad
	\subfloat[][$5$p]{%
		\includegraphics[width=5mm,height=8mm]{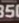}} \quad
	\subfloat[][$6$p]{%
		\includegraphics[width=5mm,height=8mm]{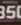}}
	\caption{Comparison of different values of padding.}
	\label{fig:padding}    
\end{figure}

Since many characters might not be perfectly segmented, containing missing parts, and as each character is relatively small, even one pixel difference between the ground truth and the prediction might impair the character's recognition. Therefore, we evaluate different padding values ($1$-$3$ pixels) in the segmented characters to achieve higher recognition rates. As Fig.~\ref{fig:padding} illustrates, the more padding pixels the more noise information is added (e.g., portions of other characters or the \gls*{lp} frame).

\vspace{-1mm}

As previously mentioned, we use two networks for character recognition. For training these networks, the characters and their labels are passed as input. For digit recognition, we removed the first four layers of the character segmentation \gls*{cnn}, since in our tests the results were similar, but with a lower computational cost. However, for letter recognition (more classes and fewer examples) we still use the entire architecture of the character segmentation \gls*{cnn}. The networks for digit and letter recognition have $75$ and $155$ filters in the last convolutional layer, respectively (see Eq.~\ref{eq:filters}).

The use of two networks allows the tuning of network parameters (e.g., input/output size) for each task. The best network sizes found in our experiments are $42$~$\times$~$26$~$\rightarrow$~$21$~$\times$~$13$ and $270$~$\times$~$80$~$\rightarrow$~$33$~$\times$~$10$ for digits and letters, respectively. 

Having knowledge of the specific \gls*{lp} country layout (e.g., the Brazilian layout), we know which characters are letters and which are digits by their position. 
We sort the segmented characters by their horizontal and vertical positions for cars and motorcycles, respectively. 
The first three characters correspond to the letters and the last four to the digits, even in cases where the \gls*{lp} is considerably tilted. 
It is worth noting that a country (e.g., USA) might have different \gls*{lp} layouts, so this approach would not be suitable in such cases.

In addition to performing the training with the characters available in the training set, we also perform data augmentation in two ways. First, we create negative images to simulate characters from other vehicle categories (as in the character segmentation stage) and then, we also check which characters can be flipped both horizontally and vertically to create new instances. Table~\ref{tab:data_augmentation} shows which characters can be flipped in each direction.
 
\begin{table}[!htb]
	\centering
	\caption{The characters that can be flipped in each direction to create new instances. We also use the numbers $0$ and $1$ as training examples for the letters O and I, respectively.}
	\label{tab:data_augmentation}
	{
		\renewcommand{\arraystretch}{1.1}
		\begin{tabular}{|c|c|}
			\hline
			Flip Direction & Characters \\ \hline
			Vertical & 0, 1, 3, 8, B, C, D, E, H, I, K, O, X \\ \hline
			Horizontal & 0, 1, 8, A, H, I, M, O, T, U, V, W, X, Y \\ \hline
			Both & 0, 1, 6(9), 8, 9(6), H, I, N, O, S, X, Z \\ \hline
		\end{tabular}
	}
\end{table}

As in the \gls*{lp} detection step, we use confidence threshold~=~$0$ and consider only the detection with the largest confidence. Hence, we ensure that a class is predicted for every segmented character.

\subsection{Temporal Redundancy}

After performing the \gls*{lp} recognition on single frames, we explore the temporal redundancy information through the union of all frames belonging to the same vehicle. Thus, the final recognition is composed of the most frequently predicted character at each \gls*{lp} position (majority vote).

Temporal information has already been explored previously in \gls*{alpr}~\cite{goncalves2016,donoser2007}. In both studies, the use of majority voting has greatly increased recognition rates.
\section{Experimental Results}
\label{sec:results}

In this section, we conduct experiments to verify the effectiveness of the proposed \gls*{alpr} system. All the experiments were performed on a NVIDIA Titan XP GPU ($3$,$840$ CUDA cores and $12$ GB of RAM) using the Darknet framework~\cite{darknet13}.

We consider as correct only the detections with \gls*{iou}~$\geq$~$0.5$. This value was chosen based on previous works~\cite{montazzolli2017,li2017towards,yuan2017}. In addition, the following parameters were used for training the networks:~$80k$ iterations (max batches) and learning rate~=~[$1$\textsuperscript{-$3$},~$1$\textsuperscript{-$4$},~$1$\textsuperscript{-$5$}] with steps at $25k$ and $35k$ iterations.

Experiments were conducted in two datasets: \gls*{ssig} and \acrshort*{dataset}. We report the results obtained by the proposed system and compare with previous work and two commercial systems\footnote{OpenALPR and Sighthound systems have Cloud APIs available at \url{https://www.openalpr.com/cloud-api.html} and \url{https://www.sighthound.com/products/cloud}, respectively. The results presented here were obtained on January, 2018.}: Sighthound~\cite{masood2017sighthound} and OpenALPR\footnote{Although it has an open-source version, the commercial version uses different algorithms for OCR trained with larger datasets to improve accuracy.}~\cite{openalpr}. According to the authors, both are robust in the detection and recognition of Brazilian \glspl*{lp}. 

It is important to emphasize that although the commercial systems were not tuned for these datasets, they use much larger private datasets, which is a great advantage especially in \gls*{dl} approaches. 

In the OpenALPR system we choose which \gls*{lp}'s style we want to detect (i.e., Brazilian) and we do not need to make any changes. On the other hand, Sighthound uses a single model for \glspl*{lp} from different countries. Therefore, we made some adjustments in its prediction so that it fits the Brazilian \glspl*{lp} format, such as swapping $0$ by O and vice versa.

\subsection{Evaluation on the \gls*{ssig} Dataset}
\label{ssig:results_ssig}

The \gls*{ssig} dataset~\cite{goncalves2016benchmark} is composed of $2$,$000$ images of $101$ vehicles with the following annotations: the position of the vehicle's \gls*{lp}, its identification (e.g., ABC-1234) and each character's position.

\begin{figure}[!htb]
	\centering
 	\includegraphics[width=0.8\columnwidth]{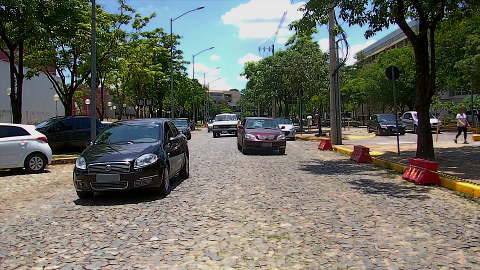}
	\caption{A sample frame of the \gls*{ssig} dataset. It should be noted that there are vehicles in the background that do not have annotations. The \glspl*{lp} were blurred due to privacy constraints.}
	\label{fig:ssig_sample}    
\end{figure}

The high resolution images ($1$,$920$ $\times$ $1$,$080$ pixels) were acquired with a static digital camera and are available in the \gls*{png} format. A sample frame of the dataset is shown in Fig.~\ref{fig:ssig_sample}.

The \gls*{ssig} dataset uses the following evaluation protocol: $40\%$ of the dataset to training, $20\%$ to validation and $40\%$ to test. According to the authors, this protocol was adopted because many character segmentation approaches do not require model estimation and a larger test set allows the reported results to be more statistically significant.

We report only the results obtained with the Fast-YOLO model in the vehicle and \gls*{lp} detection subsections, since it achieved impressive recall and precision rates in both tasks.

\subsubsection{Vehicle Detection}
Since the \gls*{ssig} dataset does not have vehicle annotations, we manually label the vehicle's bounding box on each image of the dataset. Another possible approach would be to train a vehicle detector using the large-scale CompCars dataset~\cite{yang2015}, but that way many vehicles (including those in the background) would also be detected. 

To perform the vehicle detection, we first evaluate different confidence thresholds. We started with confidence of $0.5$, however some vehicles were not detected. All $407$ vehicles in the validation set were successfully detected when the threshold was reduced to $0.25$. Based on that, we decided to use half of this value (i.e., $0.125$) in the test set to increase the chance that all vehicles are detected. With this threshold, we achieved a recall of $100\%$ and precision above $99\%$ (only $7$ false positives). 

\subsubsection{\gls*{lp} Detection}
Every vehicle in the validation set was well segmented with its \gls*{lp} completely within the predicted bounding box. Therefore, we use the vehicle patches without any margin to train the \gls*{lp} detection network. As expected, all \glspl*{lp} were correctly detected in both validation and test sets (recall and precision = $100\%$). 

\subsubsection{Character Segmentation}
A margin of $5\%$ (of the bounding box size) is required so each detected \gls*{lp} contains all its characters fully. Therefore, we double this value (i.e., $10\%$) in the test set and in the training of the character segmentation~\gls*{cnn}. 

We evaluated, in the validation set, the following confidence thresholds: $0.5$, $0.25$ and $0.1$, but the recall achieved was $99.89\%$, regardless. Therefore, we chose to use a lower threshold (i.e., $0.1$) in the test set to miss as few characters as possible. That way, we achieved $99.75\%$ ($5$,$614$/$5$,$628$) recall.

\subsubsection{Character Recognition}

The padding values that yielded the best recognition rates in the validation set were $2$ pixels for letters and $1$ pixel for digits. In addition, data augmentation with flipped characters only improved letter recognition, hampering digit recognition. We believe that a greater padding and data augmentation improve letter recognition because each class have far fewer training examples, compared to digits.

We first analyzed the results without temporal redundancy information. The proposed system achieved recognition rate of~$85.45\%$, recognizing all three letters and all four digits in~$86.32\%$ and~$98.63\%$ of the time, respectively. 

The results are greatly improved when taking advantage of temporal redundancy information. The final recognition rate is $93.53\%$, since the digits are correctly recognized in all vehicles and the letters in $93.53\%$ of them. This result is given based on the number of frames correctly recognized, thereby vehicles with more frames have greater weight in the final result.

The recognition rates accomplished by the proposed system were considerably better than those obtained in previous works ($81.8\%$~$\rightarrow$~$93.53\%$), as shown in Table \ref{tab:ssig_results}. As expected, the commercial systems have also achieved great recognition rates, but only the proposed system was able to recognize correctly at least $6$ of the $7$ characters in all \glspl*{lp}. 
This is particularly important since the \gls*{lp}'s identification can be combined with the vehicle's manufacturer/model~\cite{dlagnekov2005} or its appearance~\cite{goncalves2016} to further enhance the recognition.

\begin{table}[!htb]
	\centering
	\caption{Recognition rates obtained by the proposed \gls*{alpr} system, previous work and commercial systems in the \gls*{ssig} dataset.}
	\label{tab:ssig_results}
	{
		\renewcommand{\arraystretch}{1.1}
		\resizebox{\columnwidth}{!}{
		\begin{tabular}{@{}ccc@{}}
			\toprule
			ALPR & $\geq$ $6$ characters & All correct (vehicles)\\ \midrule
			Montazzolli and Jung~\cite{montazzolli2017} & $90.55\%$ & $63.18\%$ \\
			Sighthound~\cite{masood2017sighthound} & $89.05\%$ & $73.13\%$ \\
			Proposed & $99.38\%$ & $85.45\%$ \\
			OpenALPR~\cite{openalpr} & $92.66\%$ & $87.44\%$ \\ \midrule
			Gonçalves et al.~\cite{goncalves2016} (with redundancy) & $-$ &  $81.80\%$ ($32$/$40$) \\
			Sighthound (with redundancy) & $99.13\%$ & $89.80\%$ ($35$/$40$)\\
			OpenALPR (with redundancy) & $95.77\%$ & $93.03\%$ ($37$/$40$)\\
			\textbf{Proposed (with redundancy)} & $\textbf{100.00\%}$ & $\textbf{93.53\%}$ $\textbf{(37/40)}$ \\ \bottomrule
		\end{tabular}}
	}
\end{table}

According to our experiments, the great improvement in our system lies on separating the letter and digits recognition on two networks, so each one is tuned specifically for its task. Moreover, data augmentation was essential for letter recognition, since some classes (e.g., C, V) have less than $20$~training examples.

In Table~\ref{tab:ssig_fps}, we report the recall/accuracy rate achieved in each \gls*{alpr} stage separately, as well as the time required for the proposed system to perform each stage. The reported time is the average time spent processing all inputs in each stage, assuming that the network weights are already loaded.

\begin{table}[!htb]
	\centering
	\caption{Results obtained and the computational time required in each \gls*{alpr} stage in the \gls*{ssig} dataset. Recall stands for detection and segmentation, and Accuracy stands for recognition.}
	\label{tab:ssig_fps}
	\begin{tabular}{@{}cccc@{}}
		\toprule
		ALPR Stage & Recall/Accuracy & Time (ms) & \gls*{fps} \\ \midrule
		Vehicle Detection & $100.00\%$ & $4.0746$ & $245$ \\
		License Plate Detection & $100.00\%$ & $4.0654$ & $246$ \\
		Character Segmentation & $99.75\%$ & $1.6555$ & $604$ \\
		Character Recognition & $97.83\%$ & $1.6452$ $\times$ $7$ & $87$ \\ \midrule
		\gls*{alpr} (all correct) & $85.45\%$ & \multirow{2}{*}{$21.3119$} & \multirow{2}{*}{$47$} \\
		\gls*{alpr} (with redundancy) & $93.53\%$ & & \\ \bottomrule
	\end{tabular}
\end{table}

Since the same model is used for vehicle and \gls*{lp} detection, the time required for both stages is very similar. The same is true for character segmentation and recognition, but the latter is performed $7$ times (one time for each character). The average processing time for each frame was $21.31$ seconds, an average of $47$ \gls*{fps}.

Our system had no difficulty recognizing red \glspl*{lp}, even with less training examples. According to our experiments, this is due to the negative images used in the training of the character segmentation and recognition \glspl*{cnn}. 
Due to the agreement terms of the \gls*{ssig} dataset, we can not show qualitative results. 
Only a few \glspl*{lp} (all from the training set) can be shown for illustrations of publications.

\subsection{Evaluation on the \acrshort*{dataset} Dataset}
\label{dataset:results_ssig}

\subsubsection{Vehicle Detection}

We first evaluated the Fast-YOLO model, but the recognition rates achieved were not satisfactory. After evaluations with different confidence thresholds, the best recall rate achieved was $97.33\%$. This was expected since this dataset has greater variability in vehicle types and positions. 

\begin{figure}[!htb]
	\centering
	\includegraphics[width=0.48\columnwidth]{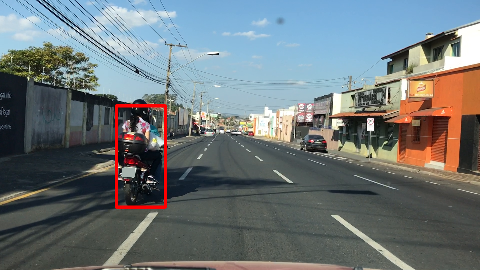}
	\includegraphics[width=0.48\columnwidth]{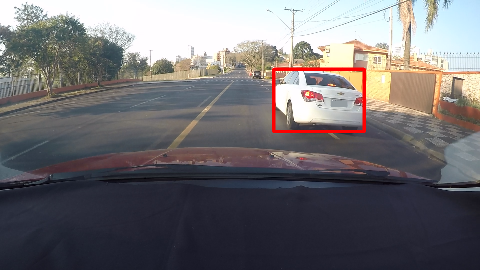}
	\caption{Examples of the detection obtained with the YOLOv2 model.}
	\label{fig:vehicle_detect_ufpr}    
\end{figure}

We chose to use the YOLOv2 model for vehicle detection, despite its higher computational cost. We evaluated several confidence thresholds, being $0.125$ the best one, as in the \gls*{ssig} dataset. The recall and precision rates achieved were $100\%$ and~$99\%$, respectively. 
Fig.~\ref{fig:vehicle_detect_ufpr} shows a motorcycle and a car detected with the YOLOv2 model.

\subsubsection{\gls*{lp} Detection}

We note that in more challenging images (usually of motorcycles), the vehicle's \gls*{lp} is not entirely within its predicted bounding box, requiring a small margin ($5\%$ in the validation set) so that the entire \gls*{lp} is completely within the predicted vehicle's bounding box. Therefore, we use a $10\%$ margin in the test set and in the training of the \gls*{lp} detection~\gls*{cnn}. 

\begin{figure}[!htb]
	\centering
	\includegraphics[width=0.9\columnwidth]{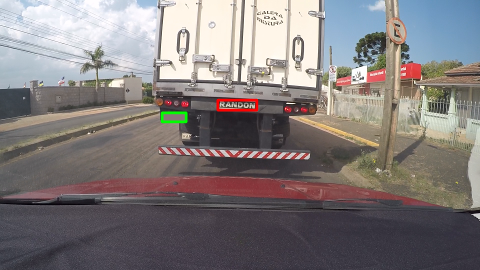}
	\caption{A sample frame from the \gls*{dataset} dataset where the actual \gls*{lp} was not predicted with the highest confidence. The predicted position and ground truth are outlined in red and green, respectively. The \gls*{lp} was blurred due to privacy constraints.}
	\label{fig:dataset_lp_fp}    
\end{figure}

The recognition rates obtained by both YOLO models were very similar (less than half a percent difference). Thus, we use the Fast-YOLO model for \gls*{lp} detection. The recall rate attained was $98.33\%$ ($1$,$770$/$1$,$800$). We were not able to detect the \gls*{lp} in just one vehicle (in its $30$ frames), because a false positive was predicted with greater confidence than the actual \gls*{lp}, as shown in Fig.~\ref{fig:dataset_lp_fp}.

We could use the character segmentation CNN to perform a post-processing in cases where more than one \gls*{lp} is detected, for example: evaluate on each detected \gls*{lp} if there are $7$ characters or consider only the \gls*{lp} where the characters’ confidence is greater. However, since the actual \gls*{lp} can be detected with very low confidence levels (i.e.,~$\leq$~$0.1$), many false negatives would have to be analyzed, increasing the overall computational cost of the system.

\subsubsection{Character Segmentation}

In the validation set, a margin of $10\%$ is required so each detected \gls*{lp} contains all its characters fully. We decided not to double the margin in the test set, as $20\%$ would add a considerable amount of noise and background in the \glspl*{lp} patches.

\begin{figure}[!htb]
	\centering
	\includegraphics[width=0.32\columnwidth]{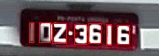}
	\includegraphics[width=0.32\columnwidth]{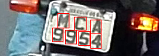}
	\includegraphics[width=0.32\columnwidth]{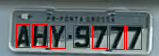}
	\caption{\glspl*{lp} from different categories properly segmented.}
	\label{fig:dataset_cs_samples}    
\end{figure}

The recall obtained was $97.59\%$ when disregarding the \glspl*{lp} not detected in the previous stage and $95.97\%$ when considering the whole test set. We accomplished better results in the \gls*{ssig} dataset, but it is worth noting that our dataset has different \glspl*{lp} types and many of them are tilted.
Fig.~\ref{fig:dataset_cs_samples} depicts some \glspl*{lp} from different categories properly segmented, even when the \gls*{lp} is tilted or in presence of shadows.

\subsubsection{Character Recognition}

The best results were obtained with $1$ pixel of padding and data augmentation, for both letters and digits. The proposed system achieved a recognition rate of $64.89\%$ when processing frames individually and $78.33\%$ ($47$/$60$ vehicles) with temporal redundancy. 

Despite the great results obtained in the previous dataset, both commercial systems did not achieve satisfactory results in the \acrshort*{dataset} dataset. 
Analyzing the results we noticed that a substantial part of the errors were in motorcycles images, highlighting this constraint in both systems.
This suggests that those systems are not so well trained for motorcycles.
OpenALPR performed better than Sighthound, attaining a recognition rate of $70\%$ when exploring temporal redundancy information. Table~\ref{tab:dataset_results} shows all results obtained in the \gls*{dataset} dataset.

\begin{table}[!htb]
	\centering
	\caption{Recognition rates obtained by the proposed \gls*{alpr} system and commercial systems in the \acrshort*{dataset} dataset.}
	\label{tab:dataset_results}
	{
		\renewcommand{\arraystretch}{1.1}
		\begin{tabular}{@{}ccc@{}}
			\toprule
			ALPR & $\geq$ $6$ characters & All correct (vehicles)\\ \midrule
			Sighthound~\cite{masood2017sighthound} & $62.50\%$ & $47.39\%$ \\ 
			OpenALPR~\cite{openalpr} & $54.72\%$ & $50.94\%$ \\
			Proposed & $87.33\%$ & $64.89\%$ \\ \midrule
			Sighthound (with redundancy) & $76.67\%$ & $56.67\%$ ($34$/$60$) \\
			OpenALPR (with redundancy) & $73.33\%$ & $70.00\%$ ($42$/$60$)\\ 
			\textbf{Proposed (with redundancy)} & $\textbf{88.33\%}$ & $\textbf{78.33\%}$ ($\textbf{47}$\textbf{/}$\textbf{60}$) \\ \bottomrule
		\end{tabular}
	}
\end{table}

We report the recall/accuracy rate achieved in each \gls*{alpr} stage separately in Table~\ref{tab:dataset_fps}, as well as the time required for the proposed system to perform each stage. The vehicle detection stage is more time-consuming in this dataset, as we use a larger \gls*{cnn} architecture (i.e., YOLOv2).

\begin{table}[!htb]
	\centering
	\caption{Results obtained and the computational time required in each stage in the \acrshort*{dataset} dataset. Recall stands for detection and segmentation, and Accuracy stands for recognition.}
	\label{tab:dataset_fps}
	\begin{tabular}{@{}cccc@{}}
		\toprule
		ALPR Stage & Recall/Accuracy & Time (ms) & \gls*{fps} \\ \midrule
		Vehicle Detection & $100.00\%$ & $11.1578$ & $90$ \\
		License Plate Detection & $98.33\%$ & $3.9292$ & $255$ \\
		Character Segmentation & $95.97\%$ & $1.6548$ & $604$ \\
		Character Recognition & $90.37\%$ & $1.6513$ $\times$ $7$ & $87$ \\ \midrule 
		\gls*{alpr} (all correct) & $64.89\%$& \multirow{2}{*}{$28.3011$} & \multirow{2}{*}{$35$} \\
		\gls*{alpr} (with redundancy) & $78.33\%$& & \\ \bottomrule
	\end{tabular}
\end{table}

It is worth noting that despite using a deeper \gls*{cnn} model in vehicle detection (i.e., YOLOv2), our system is still able to process images at $35$ \gls*{fps} (against $47$ \gls*{fps} using Fast-YOLO).  
This is sufficient for real-time usage, as commercial cameras generally record videos at $30$~\gls*{fps}.

\begin{figure}[!htb]
	\vspace{-2mm} 
	\centering
	\captionsetup[subfigure]{labelformat=empty} 
	\subfloat[][A\textcolor{red}{UC}-1056]{
		\includegraphics[width=0.31\columnwidth]{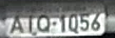}}%
    \subfloat[][BC\textcolor{red}{L}-9595]{
		\includegraphics[width=0.31\columnwidth]{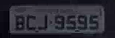}}%
    \subfloat[][AT\textcolor{red}{U}-4025]{
		\includegraphics[width=0.31\columnwidth]{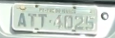}} \\[-1ex]
		
	\subfloat[][AS\textcolor{red}{D}-9\textcolor{red}{7}43]{
		\includegraphics[width=0.31\columnwidth]{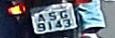}}%
    \subfloat[][AU\textcolor{red}{S}-0936]{
		\includegraphics[width=0.31\columnwidth]{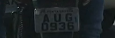}}%
    \subfloat[][AX\textcolor{red}{B}-5\textcolor{red}{4}87]{
		\includegraphics[width=0.31\columnwidth]{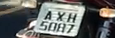}} \\[-1ex]
		
	\subfloat[][AYL-2104]{
		\includegraphics[width=0.31\columnwidth]{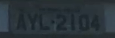}}%
    \subfloat[][AHB-1989]{
		\includegraphics[width=0.31\columnwidth]{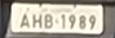}}%
    \subfloat[][AWX-9307]{
		\includegraphics[width=0.31\columnwidth]{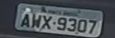}} \\[-1ex]
		
	\subfloat[][ALJ-9348]{
		\includegraphics[width=0.31\columnwidth]{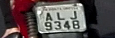}}%
    \subfloat[][AKT-8174]{
		\includegraphics[width=0.31\columnwidth]{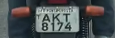}}%
    \subfloat[][MCA-9954]{
		\includegraphics[width=0.31\columnwidth]{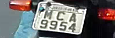}} \\[-1ex]
		
	\subfloat[][ABN-8528]{
		\includegraphics[width=0.31\columnwidth]{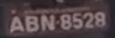}}%
    \subfloat[][AZU-3476]{
		\includegraphics[width=0.31\columnwidth]{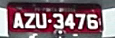}}%
    \subfloat[][AWE-4633]{
		\includegraphics[width=0.31\columnwidth]{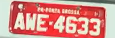}}

	\caption{Qualitative results obtained by the proposed \gls*{alpr} system in the \gls*{dataset} dataset. The first two rows shows examples of correctly detected and incorrectly recognized \glspl*{lp}, while the following rows show samples of \glspl*{lp} (from different categories) successfully recognized.}
	\label{fig:result_dataset}  
	\vspace{-2mm} 
\end{figure}

Fig.~\ref{fig:result_dataset} illustrates some of the recognition results obtained by the proposed system in the \acrshort*{dataset} dataset. It is noteworthy that our system can generalize well and correctly recognize \glspl*{lp} under different lighting conditions. 

\section{Conclusions}
\label{sec:conclusion}

In this paper, we have presented a robust real-time end-to-end \gls*{alpr} system using the state-of-the-art YOLO object detection \glspl*{cnn}.
We trained a network for each \gls*{alpr} stage, except for the character recognition where letters and digits are recognized separately (with two distinct \glspl*{cnn}).

We also introduced a public dataset for \gls*{alpr} that includes $4$,$500$ fully annotated images (with over $30$,$000$ \gls*{lp} characters) from $150$ vehicles in real-world scenarios where both vehicle and camera (inside another vehicle) are moving.
Compared to the largest Brazilian dataset (\gls*{ssig}) for this task, our dataset has more than twice the images and contains a larger variety in different aspects.

At present, the bottleneck of \gls*{alpr} systems is the character segmentation and recognition stages. 
In this sense, we performed several approaches to increase recognition rates in both stages, such as data augmentation to simulate \glspl*{lp} from other vehicle's categories and to increase characters with few instances in the training set. Although simple, these strategies were essential to accomplish outstanding results. 

Our system was capable to achieve a full recognition rate of $93.53\%$ ($85.45\%$ without temporal redundancy) in the SSIG dataset, considerably outperforming previous results ($81.8\%$ with temporal redundancy~\cite{goncalves2016} and $63.18\%$ without~\cite{montazzolli2017}) and presenting a performance slightly better than commercial systems ($93.03\%$). 
In addition, the proposed system was the only to correctly recognize at least $6$ characters in all \glspl*{lp}.

We also evaluated our proposed \gls*{alpr} system and two commercial systems as baselines on the new dataset.
The results demonstrated that the \gls*{dataset} dataset is very challenging since both commercial systems reached recognition rates below $70\%$.
Our system performed better, with recognition rate of $78.33\%$. 
However, this result is still not satisfactory for some real-world \gls*{alpr} applications.

As future work, we intend to explore new \gls*{cnn} architectures to further optimize (in terms of speed) vehicle and \gls*{lp} detection stages.
We also intend to correct the alignment of inclined \glspl*{lp} and characters in order to improve the character segmentation and recognition.
Additionally, we plan to explore the vehicle's manufacturer and model in the \gls*{alpr} pipeline as our new dataset provides such information.
Although our system was conceived and evaluated on two country-specific datasets from Brazil, we believe that the proposed \gls*{alpr} system is robust to locate vehicle, \glspl*{lp} and alphanumeric characters from any other country.
In this direction, aiming a fully robust system we just need to design a character recognition module that is independent of the \gls*{lp} layout.
\section*{Acknowledgments}
This work was supported by grants from the National Council for Scientific and Technological Development~(CNPq) (\#~428333/2016-8, \#~311053/2016-5 and \#~313423/2017-2), the Minas Gerais Research Foundation~(FAPEMIG) (APQ-00567-14 and PPM-00540-17) and the Coordination for the Improvement of Higher Education Personnel~(CAPES) (DeepEyes Project).

We thank the NVIDIA Corporation for the donation of the GeForce GTX Titan XP Pascal GPU used for this research. 

\balance
\bibliographystyle{IEEEtran}
\bibliography{IEEEabrv,bibtex}

\end{document}